\def\year{2020}\relax
\documentclass[letterpaper]{article} % DO NOT CHANGE THIS
\usepackage{aaai20}  % DO NOT CHANGE THIS
\usepackage{times}  % DO NOT CHANGE THIS
\usepackage{helvet} % DO NOT CHANGE THIS
\usepackage{courier}  % DO NOT CHANGE THIS
\usepackage[hyphens]{url}  % DO NOT CHANGE THIS
\usepackage{graphicx} % DO NOT CHANGE THIS
\usepackage{graphicx}  % DO NOT CHANGE THIS
\usepackage{textcomp}
\usepackage{color}
\usepackage{multirow}
\usepackage{booktabs}
\usepackage{array}
\usepackage{lipsum}
\usepackage{filecontents}
\usepackage{algorithm}% http://ctan.org/pkg/algorithms
\usepackage{algpseudocode}% http://ctan.org/pkg/algorithmicx
\title{Hierarchical Context Enhanced Multi-Domain Dialogue System for Multi-domain Task Completion}
\author{Jingyuan Yang,Guang Liu\thanks{Corresponding author},\\
\Large \textbf{Yuzhao Mao, Zhiwei Zhao, Weiguo Gao, Xuan Li,  Haiqin Yang, Jianping Shen}\\
\\ 
\{yangjingyuan743,liuguang230\}@pingan.com.cn \\
\{maoyuzhao258,zhaozhiwei387,gaoweiguo801,lixuan208,yanghaiqin260,shenjianping324\}@pingan.com.cn
}
 
\begin{document}

\maketitle

\begin{abstract}
Task 1 of the DSTC8-track1 challenge aims to develop an end-to-end multi-domain dialogue system to accomplish complex users’ goals under tourist information desk settings. This paper describes our submitted solution, Hierarchical Context Enhanced Dialogue System (HCEDS), for this task. The main motivation of our system is to comprehensively explore the potential of hierarchical context for sufficiently understanding complex dialogues. More specifically, we apply BERT to capture token-level information and employ the attention mechanism to capture sentence-level information. The results listed in the leaderboard show that our system achieves first place in automatic evaluation and the second place in human evaluation.
 
\end{abstract}

\section{Introduction}
Task-oriented dialogue systems aim to help users to accomplish specific tasks, e.g., booking a ticket, checking the weather. An intelligent dialogue system can significantly reduce labour expenses and improve work efficiency. Due to its promising prospect in many areas, developing intelligent dialogue systems draws much attention in both academia and industry.\\

Generally, a dialogue system consists of three components \cite{mehri2019structured}: the natural language understanding (NLU) module, the dialogue management (DM) module, and the natural language generation (NLG) module.  The NLU module is the entry to the whole system and needs to correctly understand an inputted utterance to guarantee the performance of the rest components. The DM module receives the output from the NLU module, maintains the dialog states (DS), and predicts the system actions through designed policy. The NLG module takes system actions as input to generate responses, which fulfil users’ requests.\\

Early studies usually build dialogue systems to deal with simple users’ goals within a single domain \cite{williams2013dialog,mrkvsic2016neural,mairesse2009spoken,yan2017building,wen2016network,li2017end,peng2017composite}. However, real-world dialogue systems usually need to deal with complex users’ goals spanning over multiple domains. Recent studies take much effort to explore the methods in constructing multi-domain dialogue systems \cite{ultes2017pydial,miller2017parlai}. Unlike single domain dialogue systems, a multi-domain dialogue system may encounter domain ambiguity because of slot overlapping and lacking context information \cite{rastogi2017scalable}. For example, there is no clear clue to infer the domain of the utterance ``I want free parking''. It may happen in the domain of HOTEL, ATTRACTION or any others proper domains. With sufficient context information, the domain can be easily inferred to understand users’ need.\\

To further explore the potential of dialog context, we propose a Hierarchical Context Enhanced Dialogue System (HCEDS). More specifically, we utilize token-level contextual information encoded from BERT, and sentence-level contextual information from utterance guided attention on dialogue history for better understanding user utterance. As a result, HCEDS achieves the best performance in success rate, precision, recall and f1 scores through machine evaluation, and the second place in human evaluation.\\
 
The rest of this paper is organized as follows.  First, we give a short review of  related work. Second, the task definition is described. Third, the architecture of HCEDS is reported. Next, the experimental settings are presented. Then, the experimental results and analysis are given. Last, the conclusion is delivered.

\section{Related Works}
Modular task-oriented dialogue systems achieve promising results in various scenarios. It usually consists of three components: the NLU module, the DM module, and the NLG module. In this section, we make a brief review of them.

Usually, the NLU module in mulit-domain dialog systems consists of classifiers or sequence-taggers, which try to categorize the corresponding domains, intents, and slots.  Previously proposed multi-domain NLU modules apply independent classifiers to categorize them, but ignore the  complementary information among domains \cite{mairesse2009spoken}. To address this issue, ONENET \cite{kim2017onenet} assumes that an utterance belongs to one specific domain and jointly learns three classifiers exploring the complementary information of multiple domains. However, an utterance usually belongs to multiple domains and may contain multiple intents. MILU \cite{lee2019convlab} applies multi-label classification setting and proposes an integrated classifier to classify domains and intents simultaneously. However, without clear domain information, the domain-intent-slot triplets may be in multi-turn dialogue \cite{rastogi2017scalable}. It still needs much effort to explore contextual information to enhance the domain-intent-slot categorization.

The objective of the DM module is to a dialog state tracker (DST) to track dialogue states and to choose appropriate decisions based on dialogue policy. For multi-domain DST, rnn-based models using Gated Recurrent Units (GRU) \cite{rastogi2017scalable}, Bidirectional Long Short-Term Mmemory (BiLSTM)  \cite{ramadan2018large} have been applied. Rule-based DST is another branch of active methods in the community \cite{williams2013dialog}. In terms of dialogue policy learning, they can be categorized into rule-based, supervise-based \cite{devault2011evaluation} and reinforcement learning (RL) based methods \cite{cuayahuitl2015strategic}, respectively. Recently, due to lack of enough data for learning well-set dialogue policy\cite{schatzmann2006survey}, RL based methods are intensively investigated. 

The NLG module converts system actions processed from DM to natural language. Template-based NLG \cite{lee2019convlab}  is the simplest, yet efficient approach to extract  system actions from DM. It maps input semantic symbols into tree-like or template structures and converts the intermediate structures into sentences, which respond to users’  requests. \cite{walker2002training}.Neural network-based approaches, e.g., LSTM-based structure with the Recurrent Neural Network Language Model(RNNLM) loss \cite{wen2015stochastic}, have been popularly applied for NLG \cite{wen2015semantically} .

\section{Task description}
\begin{figure}[ht]
	\centering
	\includegraphics[width=7.0cm]{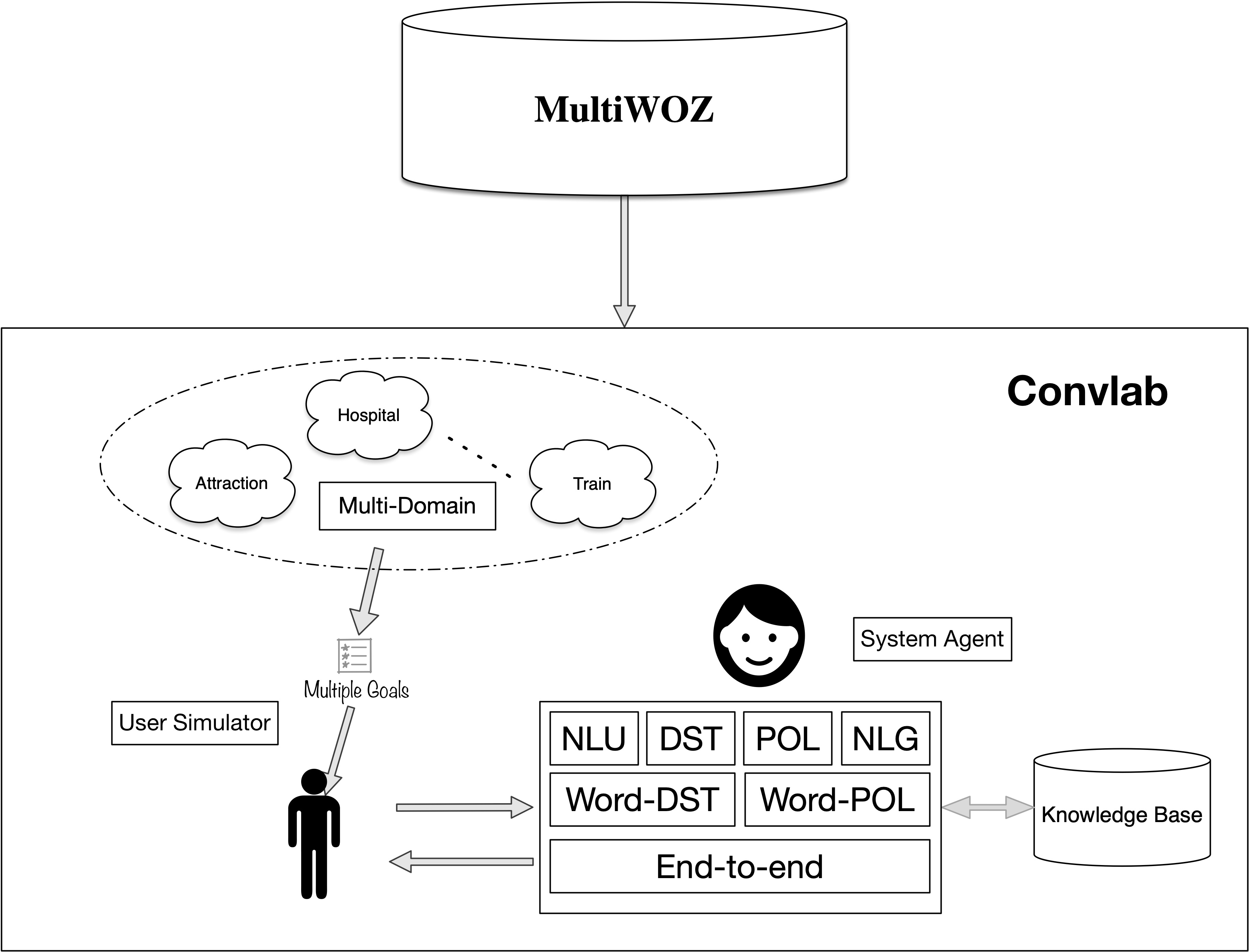}
	\caption{Task description\label{fig:1}}
\end{figure}
Task1 of DSTC8-track1 aims to encourage participants to build an end-to-end multi-domain dialogue system under the setting of the Tourist Information Desk. They provide an open-source dialog system platform, ConvLab \cite{lee2019convlab}, which provides off-the-shelf APIs for quickly setting up experiments. To fully evaluate the submitted dialogue systems, DSTC8 offers two evaluating strategies which are simulation-based evaluation and crowdworker-based evaluation.\\

As shown in Figure \ref{fig:1}, the task is based on the Multi-Domain Wizard-of-OZ (MultiWOZ) \cite{budzianowski2018multiwoz}, which is a large-scale multi-domain dialogue dataset containing seven domains related to traveling. The ConvLab provides a convenient interface and reference models for MultiWOZ dataset. The user simulator is built by obeying the static from MultiWOZ. Additional annotations for user dialog acts are further annotated for multi-domain NLU research. 

%UA:domain-intent[slot-pairs]
%Request StateRS: Request-history slot pairs
%Dialogue Historydialogue history(DH)
%Termination condition: all goals that satisfy user or reach the maximum number of rounds

\section{Hierarchical Context Enhanced Dialogue System}
\begin{figure*}
	\centering
	\includegraphics[width=16.0cm]{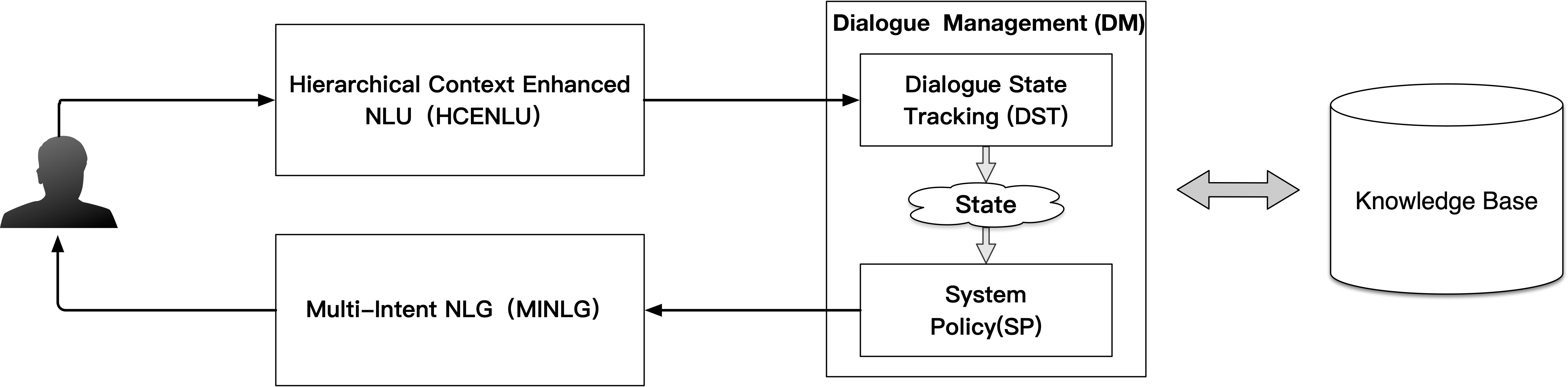}
	\caption{System architecture\label{fig:2}}
\end{figure*}

Figure \ref{fig:2} illustrates our designed Hierarchical Context Enhanced Dialogue System (HCEDS), which consists of following three modules.
\begin{equation}
HCEDS = <HCENLU, DM, MINLG>,
\end{equation}
where $HCENLU$ is our proposed NLU module (namely Hierarchical Context Enhanced NLU), $DM$ is our refined Dialogue Management module, and $MINLG$ is our proposed Multi-Intent NLG module.
\subsection{Hierarchical Context Enhanced NLU}
The HCENLU parses a user's utterance into a triplet corresponding to the value in    domain, intent and slot, respectively. For example, typical triplet lies in $\{Hotel-Request:[Price: cheap, Parking: yes]\}$. 

To resolve the problems of domain ambiguity and slot overlapping in multi-domain dialogues, we parse utterance by utilizing contextual information across different semantic levels. Different from the Hierarchical attention networks (HAN)\cite{yang2016hierarchical} for document classification,  we apply BERT\footnote{The pre-trained model and source code are from https://github.com/huggingface/transformers/tree/v0.6.1} with self-attention mechanism to capture context information within a sentence, and utilize the attention mechanism\cite{luong2015effective} to absort related context information from multi-turns dialogue hisotry according to the user's utterance. As shown in Figure 3, HCENLU contains four different layers: 1) input layers 2)token encoder layer 3) sentence encoder layer 4) output layer\\

\begin{figure}[ht]
	\centering
	\includegraphics[width=7.0cm]{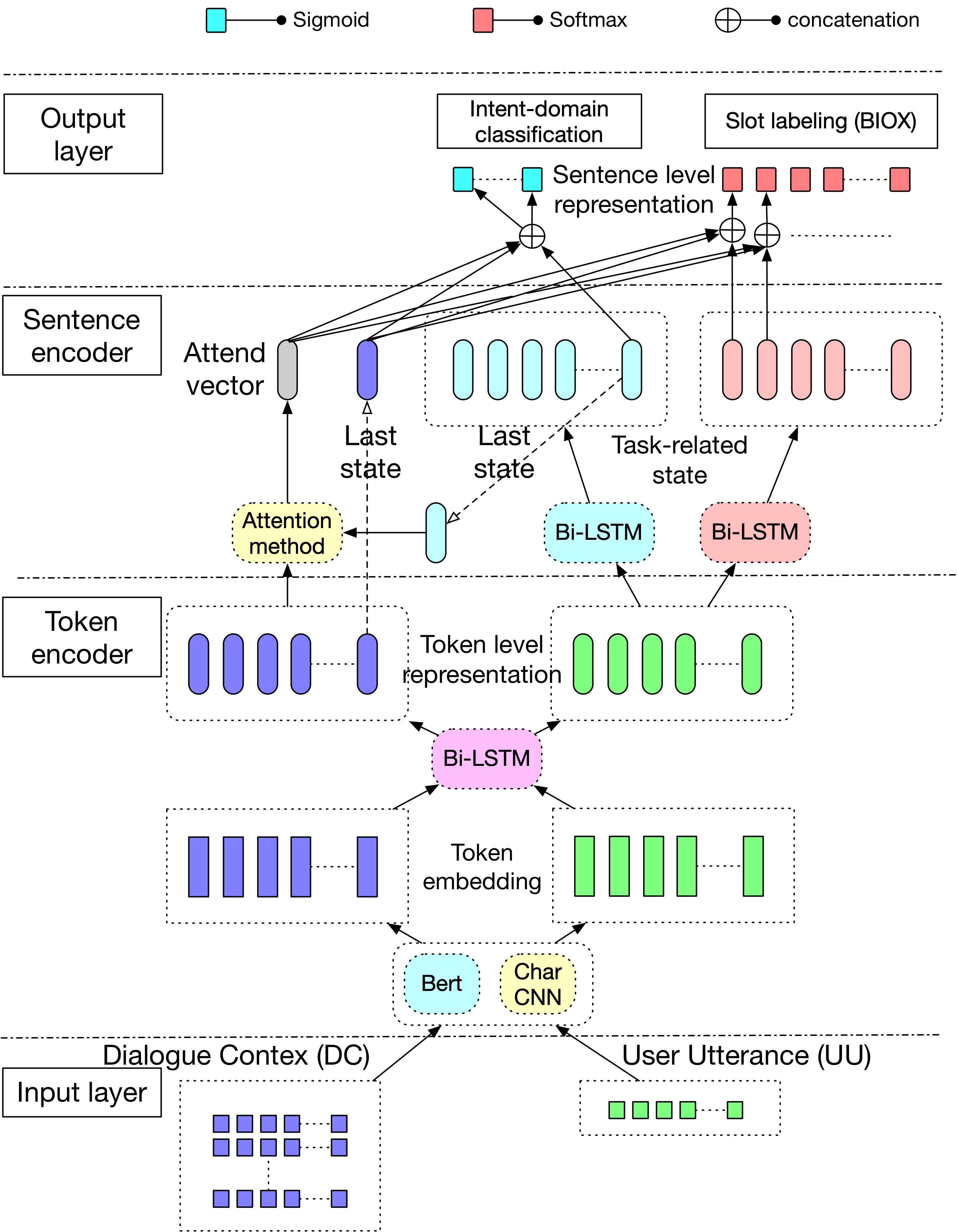}
	\caption{Illustration of HECNLU module\label{fig:3}}
\end{figure}
\textit{Input layer} has two sources of input, Users' Utterances (UU) and the Dialogue Context, namely past $w$ turn dialogues (DC). UU is a token sequence $X_{uu}={x_1, x_2, \cdots, x_m}$, where $m$ is the number of tokens of UU. The DC stacks previous $w$-turn dialogue, and is represented as a sequence of token which is $X_{dc}={x_1, x_2, \cdots, x_n}$, where $n$ is the total number of tokens of DC.\\

\textit{Token-level encoder} encodes the aforementioned two sources of sequences into contextualized token-level representation, which are transformed to token embeddings, and fed into Bi-diretional LSTM (BiLSTM), which uses two LSTM\cite{hochreiter1997long} in opsite input order, to acquire contextualized token level representation.\\

A three-step progress is carried out to get the token level representation for each input source. Firstly, the tokenized inputs are fed into a pre-trained BERT\cite{devlin2018bert} to obtain the embeddings $e^{bert}$. Secondly, each token is fed into a Char Convolutional Neural Network (CharCNN), which is a modified convolutional neural network\cite{lecun1998gradient}, to obtain a char embeddings $e^{cnn}$. Then, we concatenate two embeddings to obtain the token embedding $e^{token}$. Given an input token $x_i$, the token embedding is calculated as follows,
\begin{equation}
e^{token}_i= e^{bert}_i \oplus e^{cnn}_i.
\end{equation}
To this end, the token embeddings, UU ($e^{token}_{uu}$) and DC ($e^{token}_{dc}$), are obtained from the aforementioned two sources of sequence input. The token embeddings are fed into two independent BiLSTM separately obtaining the token-level representa-tions which is,
\begin{eqnarray}
h^{sentence}_{uu} = BiLSTM_{uu}(e^{token}_{uu}),\\
h^{sentence}_{dc} = BiLSTM_{dc}(e^{token}_{dc}),
\end{eqnarray}
where $h^{sentence}_{uu}$ and $h^{sentence}_{dc}$ are the contextualized token-level representations for UU and DC.\\

\textit{Sentence level encoder} encodes the token-level representation into sentence-level representation for domain-intent classification and tag labeling. \\
Firstly, we input the token level representations of UU into two independent BiLSTM as follows,
\begin{eqnarray}
h^{intent}_{uu} = BiLSTM(h^{sentence}_{uu}),\\
h^{tag}_{uu} = BiLSTM(h^{sentence}_{uu}).
\end{eqnarray}
Here, $h^{domain-intent}_{uu}$ is the hidden states for domain-intent classification, $h^{tag}_{uu}$ is the hidden states for tag labeling.\\ 
Secondly, we calculate the attended context vector using bilinear attention method\cite{luong2015effective}, which uses the last hidden state of $h^{domain-intent}_{uu}$ to guide the calculation of attention weights from $h^{sentence}_{dc}$, 
\begin{equation}
c^{dc} = Attention(h^{intent}_{uu},h^{sentence}_{dc}),
\end{equation}
where $c^{dc}$ is the attended context vector, the last state of $h^{intent}_{uu}$ is the quetry and  the  $h^{sentence}_{dc}$  is the key and value.\\
Finally, we concatenate three contextual vectors, $c^{dc}$ and the last hidden states of $h^{sentence}_{dc}$ and $h^{domain-intent}_{uu}$, as the sentence-level representation for domain-intent classification.
\begin{equation}
h^{intent}_{concat}=c^{dc}\oplus h^{sentence}_{dc}\oplus h^{intent}_{uu}.
\end{equation}
Here, $h^{domain-intent}_{concat}$ is the sentence-level representation for the domain-intent classification.\\
We also can get the sentence-level representation for slot labeling.
\begin{equation}
h^{tag}_{concat}=c^{dc}\oplus h^{sentence}_{dc}\oplus h^{tag}_{uu},
\end{equation}
where $h^{tag}_{concat}$ denotes the sentence-level representation for slot labeling.\\

\textit{Output layer} projects the sentence level representations into the form of domain-intent classification or slot labeling. $h^{domain-intent}_{concat}$ is fed to a multi-label classifier for domain-intent classification. For slot labeling, we apply softmax at each time-step of $h^{tag}_{concat}$ layer for tag labeling. We use BIOX tag to adapt Byte-Pair Encoding (BPE) \cite{sennrich2015neural} strategy. 

\subsection{Dialogue Management}
This module consists of two submodules,  Dialogue State Tracking (DST) and System Policy (SP).

\textit{DST} uses the default method provided by ConvLab to track dialog state\cite{williams2013dialog}. Its inputs are user actions parsed from NLU, and the dialog state $S_{t-1}$ at time ${t-1}$. Its output is the dialog state $S_t$ at time $t$, which was updated by recording user actions from time $t$ to $t-1$.\\

\textit{SP} refines the provided rule-based system policy by carefully tuning rules through bad case analysis. Its input is the dialog state $S_t$ at time $t$, and its output is the system action ${A_t}$ at time $t$, which can be represented as domain-intent-slot triplets.

\begin{table}[ht]
	\centering
	\scriptsize
	\caption{Example of Policy\label{tab:1}.}
\begin{tabular}{l|l}
\hline
\textbf{\begin{tabular}[c]{@{}l@{}}Input\\ Dialog State\\ $S_t$ \end{tabular}}  & \begin{tabular}[c]{@{}l@{}}\{\\ 'user\_action': \{'Attraction-Inform': {[}{[}'Type', 'college'{]}{]}\}, \\ 'belief\_state': \{\\     'attraction': \{\\        'book': \{'booked': {[}{]}\},\\         'semi': \{'type': 'college', 'name': '', 'area': '', 'entrance fee': ''\}\}\\      \},\\     ,...\\ \},\\ 'request\_state': \{\}, \\ 'history': {[}{[}'null', "I ' m looking for a college type attraction ."{]}{]}\\ \}\end{tabular} \\ \hline
\textbf{\begin{tabular}[c]{@{}l@{}}Ouput\\ System Action\\ $A_t$ \end{tabular}} & \begin{tabular}[c]{@{}l@{}}\{\\ 'Attraction-Recommend': {[}{[}'Name', "christ's college"{]}{]}\\ \}\end{tabular}                                                                                        \\ \hline
\end{tabular}
\end{table}

\begin{algorithm}[h] 
	\caption{Refined Rule Policy} 
	\hspace*{\algorithmicindent} \textbf{Input:} Dialog State $S_t$\\
	\hspace*{\algorithmicindent} \textbf{Output:} System Action $A_t$ \\
	\hspace*{\algorithmicindent} \textbf{Initialize:} System Action List $A = [],$ \\
	\hspace*{\algorithmicindent} Domain List $D = [hotel, restaurant, police,\\
	\hspace*{\algorithmicindent} taxi, attraction, hospital, train]$
	\begin{algorithmic}[1] 
		\For{$domain, intent, slot$ in $S_t$} 
		\State $db\_result$ = $query\_db$($domain$, $intent$, $slot$)
		\If {$intent$ is $request$}
		\If {$domain$ in $D$}
		\State$A.add$($request\_policy$($db\_result$, $domain$, $slot$))
		\EndIf
		\ElsIf {$intent$ is $inform$}
		\If {$domain$ in $D$}
		\State $A.add$($inform\_policy$($db\_result$, $domain$, $slot$))
		\EndIf
		\Else
		\State $A.add$($general\_policy$($domain$, $intent$, $slot$))
		\EndIf
		\EndFor \\
		\\
		\Return merge\_rule($A$)
		\label{code:recentEnd} 
	\end{algorithmic} 
\end{algorithm}

\subsection{Multi-intent NLG}
Multi-intent NLG (MINLG) extends the default language generating template from single-intent to multi-intent. It means that a system action may have more than one slot-value pairs. We would try to search related multi-intent templates rather than using single-intent template many times. The multi-intent templates are mined from the MultiWoz Dataset. Table.\ref{tab:1} lists an example of multi-intent templates. Obviously, multi-intent templates based NLG is able to generate more natural responses than single-intent templates based NLG.

\begin{table}[ht]
	\centering
	\scriptsize
	\caption{Comparision of single-intent templates based and multi-intent templates based NLG \label{tab:1}.}
	\begin{tabular}{|l|l|}
		\hline
		\textbf{System Action}                                                                        & \begin{tabular}[c]{@{}l@{}}\{\\ ’Attraction-inform’: \{\{’Post’,’cb21jf’\}, \\ \{’Phone’, ’01223336265’\}\}\\ \}\end{tabular} \\ \hline
		\textbf{\begin{tabular}[c]{@{}l@{}}Single-Intent Template \\ Generated Response\end{tabular}} & \begin{tabular}[c]{@{}l@{}}The attraction phone number is 01223336265.\\ The attraction postcode is cb21jf.\end{tabular}       \\ \hline
		\textbf{\begin{tabular}[c]{@{}l@{}}Multi-Intent Template \\ Generated Response\end{tabular}}  & \begin{tabular}[c]{@{}l@{}}The attraction phone number is 01223336265. \\ and its postcode is cb21jf.\end{tabular}             \\ \hline
	\end{tabular}
	
\end{table}

%%%%%%%%%%%%%%%%%%%%%%%%%%%%%%%%%%%%%%%%%%%%%

\section{Experiments}

In this section, we first outline the experimental environment and setup. Then, we discuss the quantitative results. Last, we illustrate an example of our dialogue system.

\subsection{Experimental environment}
 
\textit{Platform.} The DSTC8-track1 challenge provides ConvLab, a platform consisting of a rich set of runtime engines, for building a multi-domain end-to-end dialogue system. With a set of reusable components provided by ConvLab, approaches ranging from conventional pipeline systems to end-to-end neural models can be effortlessly developed and conveniently compared in a common platform. Our proposed HCEDS is deployed and evaluated in the ConvLab platform.

\textit{Dataset.} We mainly apply the MultiWOZ dataset to develop HCENLU. The MultiWOZ dataset consists of 10,438 dialogues, a fully-annotated collection of human-human conversations related to tourists ranging over 7 domains, i.e., attraction, hospital, police, hotel, restaurant, taxi and train. The statistic of the dataset is listed in Table.\ref{tab:2}. We follow the standard setting and split the training, validating and test as 8,438, 1,000, and 1,000, respectively.

\begin{table}[t]
	\centering
	\caption{Statistic of MultiWOZ\label{tab:2}}
	\begin{tabular}{ll}
    \toprule
		Metric                  & MultiWOZ\\
    \midrule
		Dialogues               & 8,438\\
		Total turns             & 113,556\\
		Total tokens            & 1,490,615\\
		Avg. turns per dialogue & 13.46\\
		Avg. tokens per turn    & 13.13\\
		Total unique tokens     & 23689\\
		Slots                   & 24\\
		Values                  & 4510\\
		\bottomrule
	\end{tabular}
\end{table}

\textit{Training Settings.} Letters are turned into lower case and words are encoded by BPE. The BIOX scheme is applied with BPE. We apply both Bert and CharCNN to represent utterances. The character embedding size is 16. The number of filter and the windows size for CharCNN are 128 and 3, respectively.  BiLSTMs, shared  the same model structure, are applied to obtained token-level representations, where the input size and the size of the hidden layer are 896 and 200, respectively. In addition, BiLSTMs, shared the same model structure, are applied to obtain sentence-level representations, where the input and hidden size are 400 and 200, respectively.  The dropout rate is 0.5 for all BiLSTMs. ADAM\cite{kingma2014adam} is adopted as the optimizer with the initial learning rate being 0.001 and the gradient norm being 5.0.
 
\textit{Evaluation.} Two evaluation metrics, simulation-based evaluation and crowdworker-based evaluation, are applied to evaluate the performance of each submitted dialogue systems. Besides, we design another two evaluation strategies to measure the effect of our HCENLU and the importance of other modules. The details of these four evaluation metrics are described as follows,
\begin{itemize}
	\item Simulation-based evaluation: The submitted dialogue system is evaluated automatically by computing metrics based on dialogues with a user simulator. This evaluation comprises seven metrics, which are task success rate, average reward, average dialogue turn, precision, recall, F1, and booking rate. The task success rate is the major metric.
	\item Crowdworker-based evaluation: The submitted dialogue system has to interact with a real person and evaluated by the following four metrics: task success rate, average language under-standing scores, average response appropriateness score, and average dialogue turn. The task success rate is the major metric.
    \item Effect of hierarchical context information: It is to measure the performance of each subcomponent in the NLU modules, i.e., domain-intent classification , slot tagging, and the overall performance.
    \item Ablation study: This evaluation demonstrates the effectiveness of other engineering strategies to this system. It comprises six metrics, which are task success rate, average reward, average dialogue turn, precision, recall, and F1. 
\end{itemize}

\subsection{Quantitative Results}
\subsubsection{Simulation-based evaluation} In the ConvLab platform, the user simulator randomly generates a user’s goals to test the submitted system and automatically evaluates the performance. The results in Table \ref{tab:3} show that our HCEDS achieves the highest score in task success rate and the highest F1 score (0.93), 6.9\% higher than the second ranked participant (0.87). From the results in Table \ref{tab:5}, we observe that the major gain comes from the usage of hierarchical context. By examining the details, the HCEDS obtains significantly improvement in terms of average dialogue return and average turn number, i.e., average dialogue return of 61.56, 102\% improvement from baseline, and average 7 turns per dialogue with 93.75\% of booking rate are also far advanced than that of baseline.

\begin{table*}[ht]
	\centering
	\caption{Final results of the DSTC-8 Automatic evaluation\label{tab:3}}
	\begin{tabular}{ccccccccc}
    \toprule
		Rank& \begin{tabular}[c]{@{}c@{}}Team Submission\\ ID\end{tabular} &Success Rate  &   Return    &  Turns& Precision& Recall&   F1& Book Rate\\
    \midrule
		N/A& Baseline & 63.40\% & 30.41 & 7.67 & 0.72 & 0.83 & 0.75 & 86.37\% \\
    \midrule
		10& 504569 & 52.20\% & 15.81 & 8.83  & 0.46 & 0.75 & 0.54 & 76.38\% \\
		9 & 504524 & 54.00\% & 17.15 & 9.65  & 0.66 & 0.76 & 0.69 & 72.42\% \\
		8 & 504502 & 55.20\% & 17.18 & 11.06 & 0.73 & 0.74 & 0.71 & 71.87\% \\ 
		7 & 504666 & 56.60\% & 20.14 & 9.78  & 0.68 & 0.77 & 0.7  & 58.63\% \\ 
		6 & 504529 & 58.00\% & 23.7  & 7.9   & 0.61 & 0.73 & 0.64 & 75.71\% \\ 
		5 & 504430 & 79.40\% & 49.69 & 7.59  & 0.80 & 0.89 & 0.83 & 87.02\% \\
		4 & 504641 & 80.60\% & 51.51 & 7.21  & 0.78 & 0.89 & 0.81 & 86.45\% \\
		3 & 504651 & 82.20\% & 54.09 & \textbf{6.55} & 0.71& 0.92 & 0.78& 94.56\% \\
		2 & 504563 & 88.60\% & \textbf{61.63}& 6.69 & 0.83& 0.94 & 0.87& \textbf{96.39\%} \\
    \midrule
		1 & 504429(ours)& \textbf{88.80\%}& 61.56& 7.00 & \textbf{0.92}& \textbf{0.96}& \textbf{0.93}& 93.75\% \\
		\bottomrule
	\end{tabular}
\end{table*}

\subsubsection{Crowdworker-based evaluation} Table \ref{tab:4} lists the final results, where our HCEDS attain the second place on the task success rate. Our 	HCEDS also achieves significantly better performance than the baseline in the evaluated four metrics.  More specifically, the HCEDS achieves a 65.81\% task success rate and a 16.6\% improvement over the baseline system. The system’s NLU score reached 3.538, an increase of 14.2\% compared with the baseline. The response appropriateness score is 3.632,  2.1\% higher than the baseline. The average number of dialogues was around 15, which was reduced by about 2 turns compared with 17 turns of the baseline. Overall, we have a significant improvement over the baseline in both NLU and the average rounds.
\begin{table*}[ht]
	\centering
	\caption{Final results of the DSTC-8 HUMAN evaluation\label{tab:4}}
	\begin{tabular}{cccccc}\toprule
		Rank & 
		\begin{tabular}[c]{@{}c@{}}Team Submission\\ ID\end{tabular} &
		Success Rate&
		\begin{tabular}[c]{@{}c@{}}Language \\Understanding Score\end{tabular}&
		\begin{tabular}[c]{@{}c@{}}Response \\ Appropriateness Score\end{tabular}&
		Turns\\
    \midrule
		N/A&Baseline &     56.45\%&    3.097& 3.556&  \textsl{17.543}\\
    \midrule
		10& 504502 & 23.30\% & 2.612 & 2.65 & 15.333 \\
		9 & 504666 & 25.77\% & 2.072 & 2.258 & 16.8 \\
		8 & 504582 & 36.45\% & 2.944 & 3.103 & 21.128 \\
		7 & 504529 & 43.56\% & 3.554 & 3.446 & 21.818 \\
		6 & 504569 & 54.90\% & 3.784 & 3.824 & 14.107 \\
		5 & 504641 & 62.91\% & \textsl{3.742}& 3.815 & 14.968\\
		4 & 504651 & 64.10\% & 3.547 & 3.829 & 16.906\\
		3 & 504563 & 65.09\% & 3.538 & \textsl{3.840} & \textbf{13.884}\\
		1 & 504430& \textbf{68.32\%}& \textbf{4.149} & \textbf{4.287} & 19.507\\
    \midrule
		2 & 504429(ours) & \textsl{65.81\%} & 3.538 & 3.632 & 15.481\\
		\bottomrule
	\end{tabular}
\end{table*}

\subsection{Effects of hierarchical context information}
We compare the performance of our Hierarchical Context Enhanced NLU (HCENLU) with the three official baseline models (SVMLU, ONENET, and MILU) on domain-intent classification and slot labeling. Furthermore, we analyze the effects of hierarchical context information across token-level and sentence-level to help better understand our system.

As shown in Table \ref{tab:5}, our model achieved an 85\% F1 score for the overall performance, about 2-point higher than the best baseline model, MILU. In the domain-intent classification, our model achieves 88\% of the F1 score, about 4-point higher than the MILU. In addition, we attain 83-84\% F1 score in the slot labelling with 1-2 point higher than MILU. These improvements are majorly caused by the modeling of hierarchical context information. By further analyzing the impact of context information on different levels, we observe that
\begin{itemize}
	\item We can attain 84\% F1 score, 1-point overall improvement by adding the token level context information($+b+c$) extracted by the BERT embeddings. More specifically, the F1 score of the domain-intent classification is 87\%, a 3-point improvement than the baseline, MILU. The model with CNN embedding (+b+c) obtains slightly better overall precision than the model without CNN embedding(+b).
	\item The model achieves 85\% F1 score, 1-point overall improvement by adding extra sentence level context informaiton with tag guided attention($+s4t$). More specifically, the F1 score of the tag labelling is 84\%, a 2-points improvement than token level($+b+c$) only model. The recall of domain-intent classification also attains 1-point improvement.
	\item The overall performance remains unchanged when adding extra sentence level context informaiton with intent guided attention($+s4i$). However, the domain-intent classification benefits from this modification attaining a 1-point improvement in F1 score.
	\item We take both tag guided and intent guided attention($+s4it$). However, there is no significant improvement is spotted. 
\end{itemize}

The above results indicate that the context information in both  token-level and sentence-level can  help  to  improve the performance of  domain-intent classification and slot label tagging. Besides, intent-guided attention information is help for the domain-intent classification.

\begin{table}[ht]
	\scriptsize
	\centering
	\caption{Results of the NLU models on MultiWOZ \label{tab:5} }
	\begin{tabular}{p{1.6cm}p{0.25cm}p{0.25cm}p{0.25cm}|p{0.25cm}p{0.25cm}p{0.25cm}|p{0.25cm}p{0.25cm}p{0.25cm}}
		\toprule
		\multirow{2}{*}{Model} & \multicolumn{3}{c|}{Intent} & \multicolumn{3}{c|}{Tag} & \multicolumn{3}{c}{Overall} \\
		& R & P & F                  & R & P & F               & R & P & F \\ \midrule\midrule
		SVMLU& - & - & - & - & - & - & 47\% & 68\% & 56\% \\ 
		OneNet& - & - & - & - & - & - & 58\% & 71\% & 64\% \\ 
		MILU& 86\% & 81\% & 84\% & 85\% & 80\% & 82\% & 85\% & 81\% & 83\% \\\midrule
		HCENLU\\ 
		\textit{+b}   & 88\% & 86\% & 87\% & 86\% & 80\% & 83\% & 86\% & 82\% & 84\% \\
		\textit{+b+c}   & 88\% & 86\% & 87\% & 85\% & 80\% & 82\% & 86\% & 83\% & 84\% \\
		\textit{+b+c+s4t} & 88\% & 87\% & 87\% & 86\% & 82\% & 84\% & 86\% & 84\% & 85\% \\ 
		\textit{+b+c+s4t+a4i} & 88\% & 87\% & 88\% & 85\% & 82\% & 84\% & 86\% & 84\% & 85\% \\ 
		\textit{+b+c+s4t+a4it} & 88\% & 87\% & 88\% & 86\% & 81\% & 83\% & 86\% & 83\% & 85\% \\ \midrule
		HCEDS(final)     & 88\% & 87\% & 88\% & 86\% & 82\% & 84\% & 86\% & 84\% & 85\% \\   
		\bottomrule
	\end{tabular}
\end{table}

\subsection{Ablation study}
The ablation study analyzes the effects of each module in the system in two ways. First, we replace one of the modules in the official baseline model into modules in our system to analyze the effects of individual modules in our system . Then, we replace one module from the HCEDS into a module in the baseline system to analyze the impact of each module on our system.

Table \ref{tab:6} shows the results of the first way of ablation study. The HCENLU increases the target success rate from 64\% to 68.4\%, while SP increases the target success rate to 80.2\%, a huge improvement at 16.4\%.   MINLG attains only 0.6\% improvement  and reduce the dialog length by 0.46 turns. In summary, the performance of the system is improved by incorporating the context information and gained significantly by SP.

Table \ref{tab:7} lists the ablation study results of the second investigation. If HCENLU is removed , the success rate  will drop 9.4\% while removing the SP module, the success rate drops 17\%.  By replacing  the MINLG module, the system performance will drop about 1\% while the length of the dialogue turns increases by about 0.5 rounds.

\begin{table}[ht]
	\scriptsize
	\centering
	\caption{Baseline based Ablation study on DSTC-8 Automatic evaluation\label{tab:6} }
	\begin{tabular}{lp{1.3cm}<{\centering}p{0.4cm}<{\centering}p{0.3cm}<{\centering}p{0.6cm}<{\centering}p{0.6cm}<{\centering}p{0.6cm}<{\centering}}
		\toprule
		Model & Success Rate & Return & Turns & Precision & Recall & F1 \\
		\midrule
		baseline  & 63.40\% & 30.41 & 7.67 & 72.00\% & 83.00\% & 75.00\% \\
		baseline(new)  & 64.00\% & 31.12 & 7.68 & 74.00\% & 84.00\% & 77.00\%\\
		+HCENLU   & 68.40\% & 36.25 & 7.83 & 77.00\% & 86.00\% & 79.00\% \\
		+SP &  80.20\% & 50.98 & 7.26 & 89.00\% & 91.00\% & 89.00\% \\
		+MINLG &  64.60\% & 32.31 & 7.21 & 74.00\% & 84.00\% & 77.00\% \\
		\bottomrule
	\end{tabular}
\end{table}
\begin{table}[ht]
	\scriptsize
	\centering
	\caption{HCEDS based Ablation study on DSTC-8 Automatic evaluation\label{tab:7} }
	\begin{tabular}{lp{1.3cm}<{\centering}p{0.4cm}<{\centering}p{0.3cm}<{\centering}p{0.6cm}<{\centering}p{0.6cm}<{\centering}p{0.6cm}<{\centering}}
		\toprule
		Model      & Success Rate & Return & Turns & Precision & Recall & F1 \\
    \midrule
    HCEDS      & 88.80\% &61.56 &7.00 &92.00\% &96.00\% &93.00\%\\
    HCEDS(new) & 87.60\% &60.29 &6.83 &91.00\% &94.00\% &92.00\%\\
    -HCENLU     & 78.20\% &48.5  &7.34 &88.00\% &90.00\% &88.00\%\\
    -SP     & 70.60\% &39.46 &7.26 &77.00\% &89.00\% &81.00\%\\
    -MINLG      & 86.80\% &58.85	&7.31	&92.00\% &95.00\% &93.00\%\\
		\bottomrule
	\end{tabular}
\end{table}
\subsection{Qualitative Results}

To help extensively understand the mechanism of the HCEDS, we present a success dialogue in table \ref{tab:8}. Each row represents a round of dialogue between user simulator and system. The user simulator utters utterance according to a randomly chose pre-defined user-goal and previous respond from the dialogue system. The dialogue is completed within a simulated environment. The seven turns of dialogue occur three domain-switches (turn 2-3, 3-4 and 4-5). The example shows the ability of our system to handle domain-switch in multi-domain dialogue

\begin{table}[ht]
	\centering
	\tiny
	\caption{Case study\label{tab:8} }
	\begin{tabular}{p{8cm}}
		\toprule
		\begin{tabular}[c]{@{}l@{}} User utterance: I prefer something related to museum.\\ 
			\multicolumn{1}{l}{-Inferred user action: \{'Attraction-Inform': {'Type', 'museum'}\}}  \\ 
			\multicolumn{1}{p{7.8cm}}{-System action: \{'Attraction-Inform': \{'Choice', 23\}, 'Attraction-Recommend': \{'Name', 'broughton house gallery'\}\}} \\ 
		\end{tabular} \\
		System utterance: We have 23 of those ! Anything specific you need or just a recommendation ? You would love broughton house gallery . \\
		
		\midrule
		\begin{tabular}[c]{@{}l@{}} User utterance: Yes , what are the entrance fees ? Can I get the address please ? \\
			\multicolumn{1}{l}{-Inferred user action: \{'Attraction-Request': \{'Addr', '?'\}, \{'Fee', '?'\}\}} \\ 
			\multicolumn{1}{l}{-System action: \{'Attraction-Inform': \{'Addr', '98 king street'\}, \{'Fee', 'free'\}\}} \\
		\end{tabular} \\
		System utterance: It 's located at 98 king street . Their entrance fee is free . \\
		
		\midrule
		\begin{tabular}[c]{@{}l@{}} User utterance: How about a hotel in the moderate price range ? \\
			\multicolumn{1}{l}{-Inferred user action: \{'Hotel-Inform': \{'Price', 'moderate'\}\} }\\
			\multicolumn{1}{p{7.8cm}}{-System action: \{'Hotel-Inform': \{'Choice', 18\}, 'Hotel-Recommend': \{'Name', 'a and b guest house'\}\} }\\
		\end{tabular} \\
		System utterance: There are 18 of those . Yes , I would suggest a and b guest house.  \\
		
		\midrule
		\begin{tabular}[c]{@{}l@{}} User utterance: Can I get the postcode for both of them ? I just need the address . \\
			\multicolumn{1}{p{7.8cm}}{-Inferred user action: \{'Attraction-Request': \{'Addr', '?'\}, 'Hotel-Request': \{'Addr', '?'\}, \{'Post', '?'\}\} } \\
			\multicolumn{1}{p{7.8cm}}{-System action: \{'Attraction-Inform': \{'Addr', '98 king street'\}, 'Hotel-Inform': \{'Addr', '124 tenison road'\}, \{'Post', 'cb12dp'\}\} } \\
		\end{tabular} \\
		System utterance: Their address in our system is listed as 98 king street . The address is 124 tenison road . The postal code for that hotel is cb12dp . \\
		
		\midrule
		\begin{tabular}[c]{@{}l@{}} User utterance: I just need the address . \\
			\multicolumn{1}{l}{-Inferred user action: \{'Hotel-Request': \{'Addr', '?'\}\} } \\
			\multicolumn{1}{l}{-System action: \{'Hotel-Inform': \{'Addr', '124 tenison road'\}\} } \\
		\end{tabular} \\
		System utterance: The address is 124 tenison road . \\
		
		\midrule
		\begin{tabular}[c]{@{}l@{}} User utterance: Okay I also need a taxi that will leave by 14:30 . \\
			\multicolumn{1}{l}{-Inferred user action: \{'Taxi-Inform': \{'Leave', '14:30'\}\} } \\
			\multicolumn{1}{l}{-System action: \{'Taxi-Inform': \{'Car', 'ford'\}, \{'Phone', '83307313274'\}\} } \\
		\end{tabular} \\
		System utterance: I have booked your taxi . Be expecting a ford . Their phone number is 83307313274 . \\
		
		\midrule
		\begin{tabular}[c]{@{}l@{}} User utterance: That 's all I need today . Thanks ! Bye !. \\
			\multicolumn{1}{l}{-Inferred user action: \{'exit dialogue': True\} } \\
			\multicolumn{1}{l}{-System action: \{'close session': True\} } \\
		\end{tabular} \\
		System utterance: - \\
		\bottomrule
	\end{tabular}
\end{table}

\section{Conclusion}
In this paper, we describe our submitted Hierarchical Context Enhanced Dialogue System (HCEDS), a modular multi-domain dialogue system, for Task1 of the DSTC8-track1 challenge. Our proposed HCEDS explores the potential of hierarchical contextual information from a multi-domain dialogue system.  The NLU module gains better understanding of utterances by modeling token-level and sentence-level context information, which significantly improves the performance of domain-intent classification and slot labelling on the MultiWOZ dataset, and there has been a significant accuracy raise lead by the engineering improvement in policy. It is worth noting that the HCEDS achieves the best performance in the automatic evaluation. Our system gets second place in the manual evaluation. This paper shed some lights in exploring the potential of contextual information for multi-domain dialogue systems.

%\bibliography{mybib2}

\bibliographystyle{aaai}

\end{document}